\RequirePackage{fix-cm}
\documentclass[smallextended]{svjour3}       
\smartqed  
\usepackage{graphicx}
\usepackage{xcolor}
\usepackage{hyperref}
\usepackage{gensymb}
\usepackage{amsmath}
\usepackage{caption}
\usepackage{array}
\usepackage{tabularx}
\usepackage{subcaption}
\usepackage{color, colortbl}
\usepackage{makecell}

\usepackage{graphics} 
\usepackage{checkend}	
\usepackage{epsfig}

\newcolumntype{Y}{>{\centering\arraybackslash}X}

\definecolor{ortable}{rgb}{0.5,0.0,0.0} 
\definecolor{psc}{rgb}{0.0,0.5,0.0} 
\definecolor{vsc}{rgb}{0.5,0.5,0.0} 
\definecolor{human}{rgb}{0.0,0.0,0.5} 
\definecolor{ceilinglight}{rgb}{0.5,0.0,0.5} 
\definecolor{floor}{rgb}{0.0,0.5,0.5} 
\definecolor{mayostand}{rgb}{0.5,0.5,0.5} 
\definecolor{table}{rgb}{0.25,0.0,0.0} 
\definecolor{chair}{rgb}{0.75,0.5,0.0} 
\definecolor{wall}{rgb}{0.25,0.5,0.0} 
\definecolor{anesthesiacart}{rgb}{0.75,0.5,0.0} 
\definecolor{cannula}{rgb}{0.25,0.0,0.5} 

\begin{document}
\renewcommand{\sectionautorefname}{Sect.} 
\renewcommand{\subsectionautorefname}{Sect.} 
\renewcommand{\figureautorefname}{Fig.}

\title{A Robotic 3D Perception System for Operating Room Environment Awareness
}
\subtitle{ \textit{perceptual daVinci}}


\author{Zhaoshuo Li         \and
        Amirreza Shaban \and
        Jean-Gabriel Simard \and
        Dinesh Rabindran \and
        Simon DiMaio  \and 
        Omid Mohareri
}


\institute{Zhaoshuo Li \at
             LCSR, Johns Hopkins University, Baltimore, United States \\
              \email{zli122@jhu.edu}           
\and
Amirreza Shaban, \at
Georgia Institute of Technology, Atlanta, United States \\
\email{amirreza@gatech.edu} 
\and
 Jean-Gabriel Simard, \at
MILA, Université de Montréal, Montreal, Canada \\
\email{simardjg@mila.quebec}
\and
Dinesh Rabindran, Simon DiMaio, Omid Mohareri \at
Intuitive Surgical Inc., Sunnyvale, United States \\
\email{omid.mohareri@intusurg.com}
}

\date{Received: date / Accepted: date}

\maketitle

\begin{abstract}
\textit{Purpose:} We describe a 3D multi-view perception system for the da Vinci surgical system to enable Operating room (OR) scene understanding and context awareness. \textit{Methods:} Our proposed system is comprised of four Time-of-Flight (ToF) cameras rigidly attached to strategic locations on the da \textcolor{black}{Vinci} Xi patient side cart (PSC). The cameras are registered to the robot’s kinematic chain by performing a one-time calibration routine and therefore, information from all cameras can be fused and represented in one common coordinate frame. Based on this architecture, a multi-view 3D scene semantic segmentation algorithm is created to enable recognition of common and salient objects/equipment and surgical activities in a da Vinci OR. Our proposed 3D semantic segmentation method has been trained and validated on a novel densely annotated dataset that has been captured from clinical scenarios. \textit{Results:} \textcolor{black}{The results show that our proposed architecture has acceptable registration error ($3.3\% \pm 1.4\%$ of object-camera distances) and can robustly improve scene segmentation performance (mean Intersection Over Union - mIOU) for less frequently appearing classes ($\ge 0.013$)} compared to a single-view method. \textit{Conclusion:} We present the first dynamic multi-view perception system with a novel segmentation architecture, which can be used as a building block technology for applications such as surgical workflow analysis, automation of surgical sub-tasks and advanced guidance systems.
\end{abstract}

\section{Introduction}
\label{sec:intro}
Operating room (OR) is a complex environment, where many factors together determine the surgical outcome \cite{hull2012impact}. While the performance depends highly on the technical skills of the surgeon and OR staff, it is also influenced directly by the non-technical attributes, such as situational awareness and performance feedback. 
Thus, it is important to analyze the surgical workflow \textcolor{black}{retrospectively} or in real-time to facilitate the best patient care. Further, OR activity analysis can be used towards OR scheduling for better resource allocation and automated transcription of surgical process. However, the current most adopted OR workflow analysis approach is auditing, where the observers stay inside the OR throughout the surgery \cite{divatia2015can}. This method is highly unscalable, 
and there is a strong motivation for more efficient data collection and automated surgical activity analysis systems. 

Previously, different types of cameras have been installed on fixed locations inside the OR (e.g. ceiling mounted) to monitor and record activities \cite{twinanda2015data,liu20183d}. With the introduction of robotic-assisted surgery, robots such as the da Vinci Xi Surgical System (Intuitive Surgical Inc., Sunnyvale, US), become a potential platform for mounting the cameras. The da Vinci system consists of surgeon side console (SSC), patient side cart (PSC) and vision system cart (VSC). Given that most of the activities center around the PSC robot in a da Vinci OR, it is of particular interest for installing the cameras. Furthermore,  it can enable a dynamic multi-view sensor architecture to achieve robust and improved scene understanding.  

In order to identify or analyze the OR activities, it is important to first understand the semantics of the OR scene. With the recent advancement of deep neural networks, semantic segmentation on a single image has achieved higher than ever accuracy \cite{zhou2019review}. However, it is  still an open research how to fuse predictions from multiple views observing a common scene into a unified and improved one. The advantages of having multiple cameras include improving the robustness against occlusion and providing sensor redundancy. Furthermore, it is intuitive that when multiple cameras are looking at the same scene, the combined semantic understanding should improve as a whole. This has been previously validated by researches on both indoor scene \cite{ma2017multi} and medical image \cite{mortazi2017cardiacnet} segmentation.  

Therefore, in this work, a multi-view 3D perception system has been introduced to supplement the PSC robot. Four time-of-fight (ToF) cameras are mounted on the robot at strategic locations to maximize the coverage. Both the 2D intensity and the corresponding 3D point cloud data generated by the ToF cameras are being used in our framework for multi-view semantic segmentation. The cameras are registered to the kinematics chain of the robot via a one-time calibration process using a customized calibration fixture and rigid mountings, thus allowing the sensor package to be mounted on any Xi system without the need for re-calibration. 
The information (pixel labels) from all cameras can be fused together in the robot base coordinate frame by a novel learning-based multi-view projection and merging (MVPM) technique. We demonstrate that the MVPM 
improves the performance when compared to single view results or other 3D fusion techniques that require extensive manual parameter tuning. The contributions of this paper can be summarized as following:
\begin{itemize}
    \item Introduction of a dynamic multi-camera 3D perception system for operating room scene understanding.
    \item A novel multi-view semantic segmentation fusion algorithm, which shares and combines the label confidence via learning. 
    \item \textcolor{black}{A} one-time calibration method for robotic multi-view perception systems, which can scale the data acquisition process.  
    \item Introduction of new densely annotated single and multi-view datasets for scene parsing and semantic segmentation in robotic OR environments. 
\end{itemize}

\section{Related Work}
\label{sec:related_work}
\subsection{Setup for Surgical Activity Monitoring in the OR}
To the best of our knowledge, there is no prior work for semantic segmentation inside the OR. Instead, a review of the systems for surgical workflow analysis is presented.

The least involved setup is to analyze the endoscopic videos, which is already part of the surgical flow. \cite{dergachyova2016automatic} used surgical tool segmentation for phase detection and skill analysis. Sensor-based tracking, specifically where sensors are mounted on the surgical instruments or worn by staff, is also a common approach. RFID tags, accelerometers \cite{meissner2014sensor} and eye tracker\cite{james2007eye} has been used to record information.

Further, cameras can be mounted inside the clinical environment to observe surgical activities. In \cite{twinanda2015data}, multiple RGB-D cameras are mounted on the ceiling for X-ray-based procedures. In \cite{liu20183d}, depth-only cameras are mounted in an ICU for patient monitoring. However, the setups may not be scalable as it highly depends on the specifications of the OR. Moreover, the rigid setups may restrict extension to applications such as \textcolor{black}{intelligent setup and docking, where robot-centric sensor configurations are more advantageous}. Compared to the previous two works, this paper uses similar information sources, but installed on the da Vinci surgical systems, which removes the dependency on the OR and can enable a dynamic multi-sensor architecture.

\subsection{Multi-view Semantic Segmentation Fusion}
Lacking a large 3D point cloud dataset with a comparable size of ImageNet 
(2D RGB image dataset), 3D segmentation still suffers from a good backbone model. 
Moreover, the sparsity of the 3D information in most of the currently available depth cameras, poses challenges for geometric feature extraction. Thus, many previous works leverage dense 2D descriptors to perform tasks and back-project the 2D information to 3D. 

To the best our knowledge, the first work that presented this idea is \cite{su2015multi}, where 3D shapes are projected into different 2D views for classification purpose. The feature descriptors are extracted from 2D images, aggregated by element-wise max-pooling, and finally used to classify the 3D object. In \cite{kalogerakis20173d}, this concept is extended to multi-view segmentation, where 3D objects are projected to different view points. Segmentation results are obtained from each view and back-projected to the mesh model. The surface-based Conditional Random Field (CRF) is followed to merge the segmentation result.

Other works focus on multi-view semantic segmentation in the temporal domain, i.e. fusing frames from a single moving camera. In \cite{ma2017multi}, frames are warped to a single key frame. The segmentation prediction is generated from max-pooling among the predictions from all frames. \cite{mccormac2017semanticfusion} used deep learning for 2D segmentation and dense 3D CRF for label fusion. In all of the above cases, SLAM is required to associate adjacent frames temporally. Recently, \cite{dai20183dmv} proposed a joint 2D-3D segmentation approach where 2D descriptors are back-projected to 3D voxel grid, where 2D and 3D descriptors are combined via 3D convolution on a regular grid to infer the segmentation.

No prior work on multi-camera semantic segmentation is found. Thus, we proposed a multi-view projection and merging (MVPM) method for multi-view fusion. We compared our work with dense CRF \cite{krahenbuhl2011efficient}, which is the most commonly used technique to enforce consistency by works mentioned above.

\section{Methodology}
\label{sec:methodology}
\subsection{Setup}
\label{sec:camera_and_mounting}
\begin{figure}[htpb]
    \centering
    \parbox{0.35\linewidth}{
    \centering
        \begin{subfigure}[p]{.45\linewidth}
                \includegraphics[width=\textwidth]{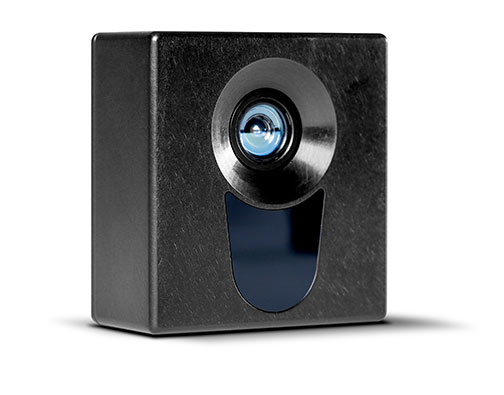}\caption{}\label{fig:camera}
        \end{subfigure}
         \begin{subfigure}[p]{.35\linewidth}
                \includegraphics[width=\textwidth]{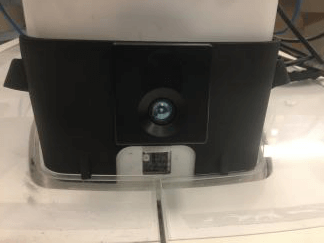}\caption{}\label{fig:base_mount}
        \end{subfigure}\\
        \begin{subfigure}[p]{.35\linewidth}
                \includegraphics[width=\textwidth]{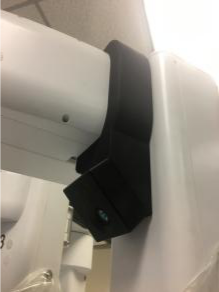}\caption{}\label{fig:usm_mount}
        \end{subfigure}
        \begin{subfigure}[p]{.35\linewidth}
                \includegraphics[width=\textwidth]{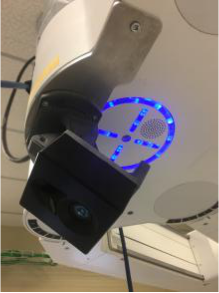}\caption{}\label{fig:op_mount}
        \end{subfigure}
    }
    \hspace{1pt}
    \begin{subfigure}{.35\linewidth}
        \centering
        \includegraphics[width=\linewidth]{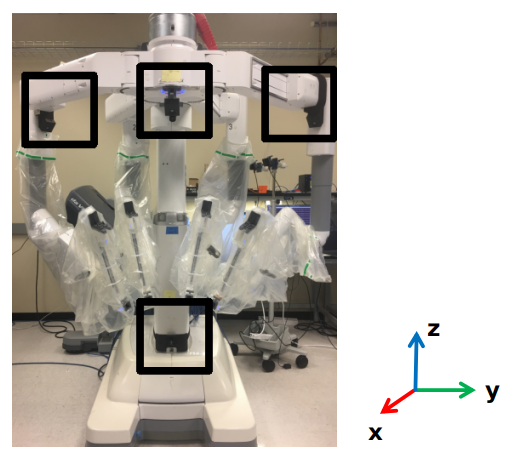}
        \caption{}
        \label{fig:setup}
    \end{subfigure}
    \caption{The setup of the proposed robotic 3D perception system, (a) the ToF camera used, (b-d) mountings for BASE, USM and OP cameras, (e) the PSC robot with ToF cameras attached (in black rectangles). OP (shown in top-center), USM1 (shown in top-left), USM4 (shown in top-right), BASE (shown in bottom). The PSC robot can only rotate around its Z axis, where the coordinate frame is shown on the right.}
\end{figure}

The physical setup of the 3D perception system is shown in \autoref{fig:setup}. There are four Pico Monstar ToF cameras (\autoref{fig:camera}) (PMDTec, Siegen, Germany) installed. The camera generates a total of $352\times287$ data points. Each data point contains two types of sensor information: the intensity and depth value. The error of the depth data is $1\%$ of the object-to-camera distance. The depth value can be further converted to a 3D point cloud expressed in the camera coordinate frame by using the provided camera intrinsics.  
The cameras are attached on the da Vinci Xi PSC robot, named respectively as Orienting Platform (OP), Universal Setup Manipulator 1 (USM1), USM4, and BASE based on their mounting locations. 

Three types of mechanical mountings are designed to provide precise camera locations on the robot without any modifications to the system. The USM1 and USM4 cameras are clamped onto the setup structure as shown in \autoref{fig:usm_mount}. The OP camera is attached to the screw holes on the orienting platform shown in \autoref{fig:op_mount}. The BASE camera is mounted on a 3D printed part that can rigidly attach to the column. The vertical position of the BASE camera is determined by the base of PSC, as shown in \autoref{fig:base_mount}. 

\subsection{Camera-to-Robot Calibration}
\label{sec:calibration}
\begin{figure}[h]
    \centering
    \parbox{0.3\textwidth}{
    \centering
        \begin{subfigure}{.35\linewidth}
                \includegraphics[width=\textwidth]{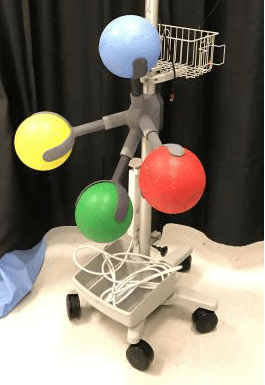}\caption{}\label{fig:fixture}
        \end{subfigure}
        \begin{subfigure}{.35\linewidth}
                \includegraphics[width=\textwidth]{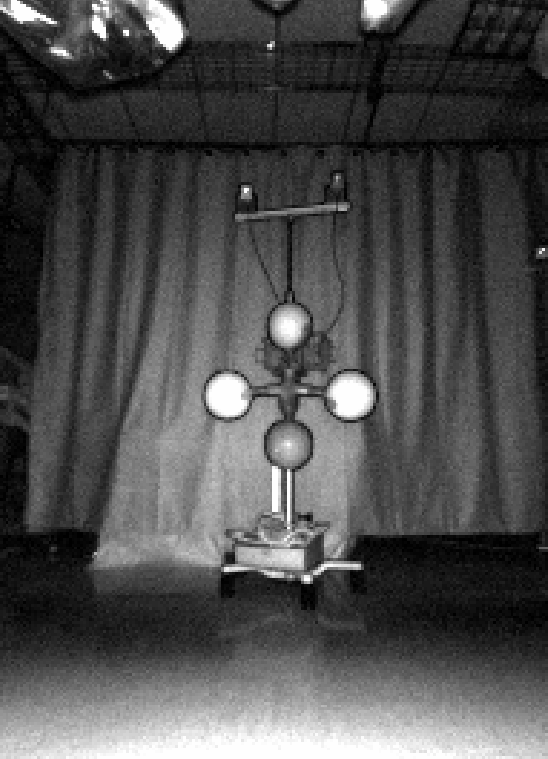}\caption{}
        \end{subfigure}\\
        \begin{subfigure}{.35\linewidth}
                \includegraphics[width=\textwidth]{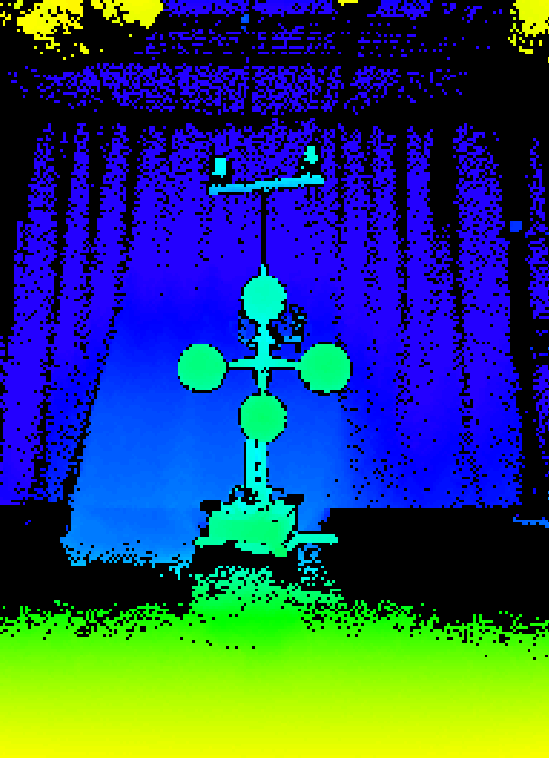}\caption{}
        \end{subfigure}
        \begin{subfigure}{.35\linewidth}
                \includegraphics[width=\textwidth]{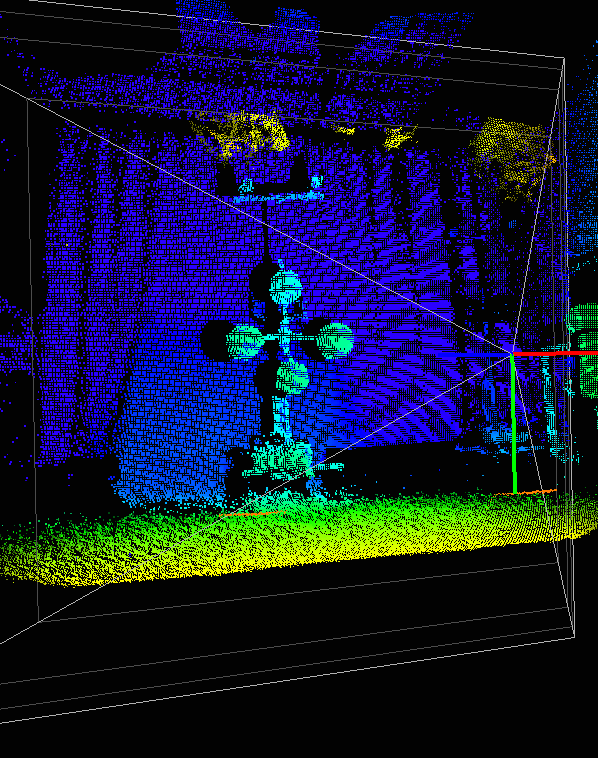}\caption{}
        \end{subfigure}
    }
    \hspace{1pt}
    \begin{subfigure}{.6\linewidth}
        \centering
        \includegraphics[width=\linewidth]{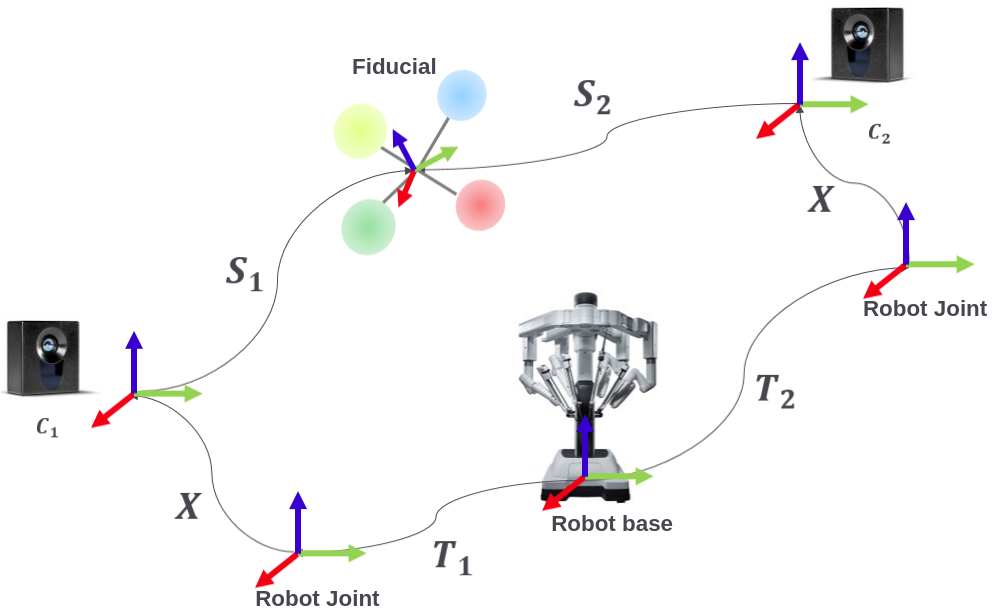}\caption{}\label{fig:calibration_process}
    \end{subfigure}
    \caption{(a) Calibration fixture, (b) ToF intensity image, (c) Depth image, (e) 3D point cloud, (f) a diagram demonstrating the calibration process.}
    \label{fig:fiducial_setup}
\end{figure}

Since the cameras are mounted on the moving components of the PSC robot to maximize the coverage, the camera locations need to be calibrated with respect to the kinematics chain of the PSC to express the surrounding objects in the PSC base frame. 
As the ToF camera generates the intensity values only using the magnitude of the infra-red (IR) light, checkerboards cannot be used for calibration since the grids are not visible. Also, since the 3D information can be directly inferred from the depth, it is more advantageous to use the 3D information for calibration directly. For this purpose, a calibration fixture has been designed as shown in \autoref{fig:fixture}. Four colored spheres of diameter approximately 20cm are rigidly attached on the fixture non-coplanarly, where different colours are used for correspondence. The 2D intensity data, depth data and 3D reconstruction of the fixture are shown in \autoref{fig:fiducial_setup}b-d. 

To calibrate the camera locations, a modified version of hand-eye calibration pipeline is used (\autoref{fig:calibration_process}). The PSC robot (denoted as robot) and calibration fixture (denotes as the colored fiducials) are kept static relative to each other. The camera ($C_1$), is mounted at a location with unknown transformation ($X$) with respect to a joint in the robot, with a known forward kinematics ($T_1$). To determine the location of the fiducials in the camera coordinate frame, a sphere fitting algorithm is used to estimate the centers of the spheres, thus best approximating the location of the fiducials. Since the four fiducials are static relative to each other, they can define a local coordinate frame. The transformation from the camera frame to the local frame of the fixture is denoted as $S_1$. The goal of the calibration process is to recover $X$ by using the above information. 

The camera is then moved to another location ($C_2$) by the robot, with a new forward kinematics ($T_2$) and new fiducial locations ($S_2$). Thus, we obtain
\begin{align}
    AX&=XB \label{eqn:hand_eye}
\end{align}
where $A=T_1^{-1}T_2$ is the relative movement of the joint, and $B=S_1S_2^{-1}$ is the inverse relative movement of the fiducials in the camera frame. However, the PSC robot is only capable of rotation around Z-axis as shown in \autoref{fig:setup}. It has been shown in \cite{andreff1999line} that we can fully recover the rotation $R_X$, while only  partial of the translation as
\begin{align}
    t_X(\alpha) = t_\bot + \alpha n_z
\end{align}
where $n_z$ is the Z-axis of the robot, $t_\bot$ is translation along the plane with normal being $n_z$, and $\alpha$ is any scalar. 

Since the BASE camera is static relative to the PSC base, the above calibration process does not apply. To calibrate the transformation from robot base to the BASE camera, 
the calibration result of OP ($X_{op}$) is used. The transformation from robot to fixture $T_{robot}^{fixture}$ can be found as
\begin{align*}
    T_{robot}^{fixture} &= X_{base}S_{base} = T_{op}X_{op}S_{op} \\
    X_{base} &= T_{op}XS_{op}S_{base}^{-1}
\end{align*}
where $X_{base}$ is the target unknown BASE camera calibration, $T_{op}$ is from the robot kinematics and $S_{base}S_{op}^{-1}$ can be estimated as before. Thus, $X_{base}$ can be solved via least squares. 

To solve for the last degree of freedom, an ICP procedure is followed by using the VSC as calibration fixture. The previous calibration result serves as a warm start for ICP algorithm. An example calibration result can be found in Appendix A.

After camera is calibrated using the above calibration process, along with the repeatable mounting described in \autoref{sec:camera_and_mounting}, the sensor package can be installed on any da Vinci Xi systems since the robots are manufactured with the same process and the mounting locations are deterministic.

\subsection{Dataset}
\label{sec:datset}
A large annotated dataset has been collected for the purpose of training deep learning models. The data has been collected in a clinical development lab, where different \textcolor{black}{robot-assisted} laparoscopic procedures are simulated, and 
videos are taken by the ToF cameras. The salient frames are then extracted from the videos, i.e. frames with significant enough differences. \textcolor{black}{The salient frame extraction process can be found in the Appendix B.} 

\begin{figure}
    \centering
    \parbox{\linewidth}{
        \centering
         \begin{subfigure}{0.22\linewidth}
            \centering
            \includegraphics[width=\linewidth]{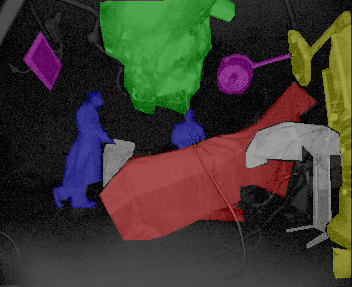} \caption{}
        \end{subfigure}
        \hspace{1pt}
        \begin{subfigure}{0.22\linewidth}
            \centering
            \includegraphics[width=\linewidth]{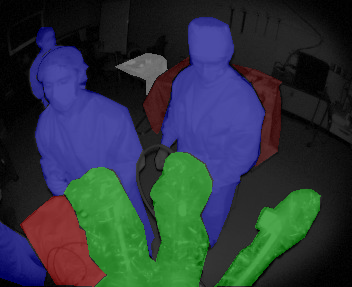} \caption{}
        \end{subfigure}
        \hspace{1pt}
        \begin{subfigure}{0.22\linewidth}
            \centering
            \includegraphics[width=\linewidth]{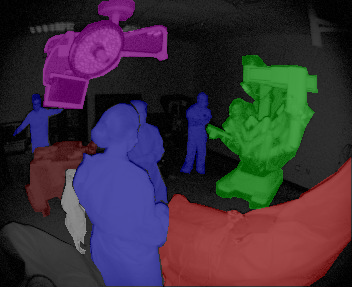} \caption{}
        \end{subfigure}
    }
    
    \parbox{\linewidth}{
        \centering
         \begin{subfigure}{0.22\linewidth}
            \centering
            \includegraphics[width=\linewidth]{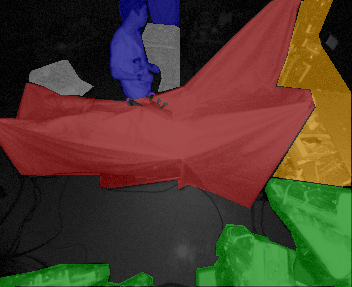} \caption{}
        \end{subfigure}
        \hspace{1pt}
        \begin{subfigure}{0.22\linewidth}
            \centering
            \includegraphics[width=\linewidth]{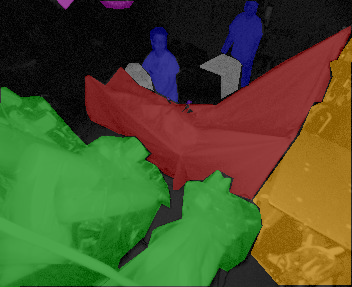} \caption{}
        \end{subfigure}
        \hspace{1pt}
        \begin{subfigure}{0.22\linewidth}
            \centering
            \includegraphics[width=\linewidth]{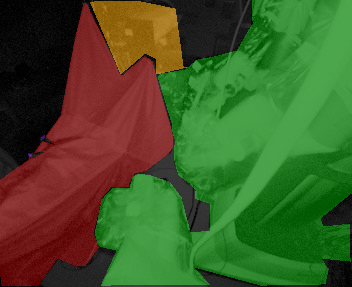} \caption{}
        \end{subfigure}
        \hspace{1pt}
        \begin{subfigure}{0.22\linewidth}
            \centering
            \includegraphics[width=\linewidth]{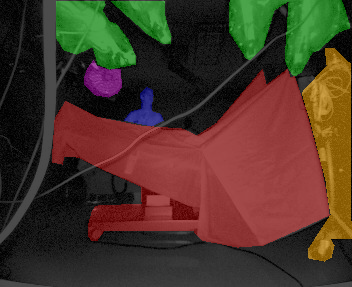} \caption{}
        \end{subfigure}
    }
    
    \begin{subfigure}{0.4\linewidth}
        \centering
        \includegraphics[width=\linewidth]{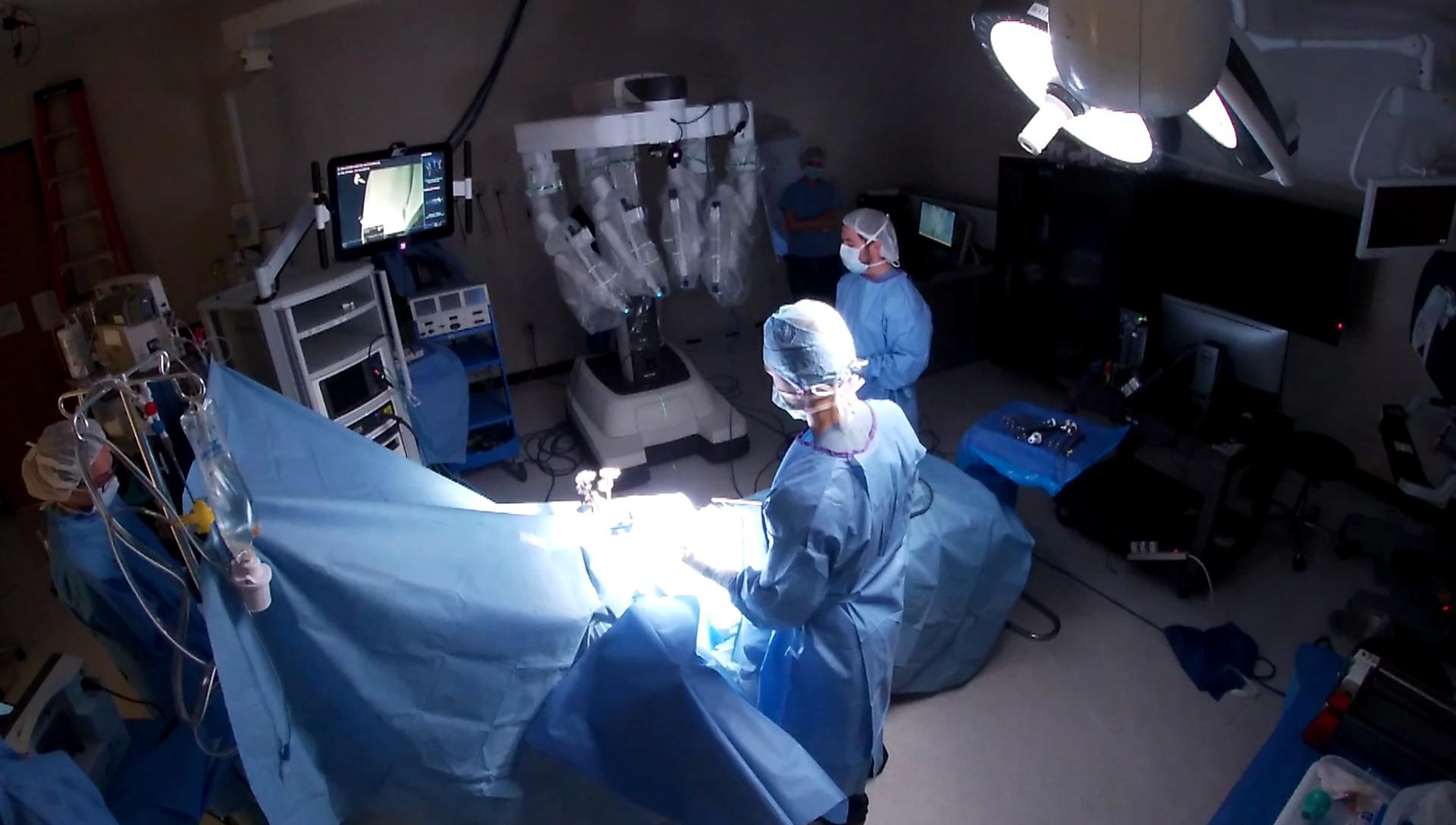} \caption{}
    \end{subfigure}

    \caption{Overlays of segmentation labels and ToF intensity images, (a-c) different view points from single-view dataset, (d-g) OP, USM1, USM4, BASE camera view points from multi-view dataset, (h) RGB image of the OR setup.}
    \label{fig:intensity_overlay}
\end{figure}

There are two portions of the dataset - single-view and multi-view. The single-view dataset consists of 7980 images. The data is captured by attaching the ToF cameras on the PSC and the VSC. Examples of the single-view data is shown in \autoref{fig:intensity_overlay}. The multi-view dataset consists of 400 images. The setup is the same as the proposed multi-view perception system, where 100 images from each of the OP, USM1, USM4, and BASE cameras are collected. Captured images and robot kinematics data are time synchronized. The example images are shown in \autoref{fig:intensity_overlay}. \textcolor{black}{The color code and pixel frequency for the 8 classes in the dataset (excluding background) is shown in \autoref{tab:data_classes}}.

\begin{table}[htpb]
    \centering
    \caption[justification=centering]{Color code and pixel frequency (in parenthesis) for each class}
    \label{tab:data_classes}       
    \begin{tabularx}{\textwidth}{YYYY}
    \hline\noalign{\smallskip}
    OR Table ($43.29\%$) & PSC ($41.13\%$) & VSC ($0.60\%$) & Human ($5.42\%$)\\
    \cellcolor{ortable} & \cellcolor{psc} & \cellcolor{vsc} & \cellcolor{human} \\
    \makecell{Ceiling Light\\($0.35\%$)} & \makecell{Mayo Stand\\($1.76\%$)} & \makecell{Sterile Table\\($2.99\%$)} & \makecell{Anesthesia Cart\\($4.42\%$)} \\
    \cellcolor{ceilinglight} & \cellcolor{mayostand} & \cellcolor{table}& \cellcolor{anesthesiacart} \\
    \noalign{\smallskip}\hline
\end{tabularx}
\end{table}

\subsection{Multi-view Semantic Segmentation}
\label{sec:semseg}
To share the segmentation label confidence between the cameras, the multi-view projection and merging (MVPM) is proposed. 
A single-view segmentation model, DeepLab V3+ \cite{chen2018encoder}, is first trained to serve as the baseline. This model outputs per-class confidence values for each pixel in the image. The outputs of the network are 4 probability tensors (one for each camera) with size $C\times352\times287$, where $C$ is the the number of classes. 

In the following, OP camera is used to explain the concept w.l.t.g. By using the relative transformation between OP camera and other cameras as well as the intrinsics of OP camera, the probabilities from USM1, USM4 and BASE cameras can be projected to the OP camera plane, resulting three times more information with pixel correspondence (Appendix C for projection examples). 
To approximate the CRF algorithm, depth image is also being projected, resulting a channel size of $4\times(C+1)$. After the projection operation, the probability and depth values are sorted by the merging module such that the confidence and depth values of the OP camera is always in the first $C+1$ channels. The rest of the cameras are sorted in the following order: USM1, USM4, BASE. This allows the merging module to always see a deterministic information from the cameras. 

Instead of max-pool the highest probability, an hourglass architecture similar to \cite{ronneberger2015u} is used to combine the $4\times(C+1)$ into $C$ channels. The probabilities in different classes are first compressed into low-level embeddings, and the embeddings are then converted to a segmentation map by the decoding process. The final class is assigned by the class with maximum likelihood. 

\begin{figure}[htpb]
    \centering
    \includegraphics[width=0.8\linewidth]{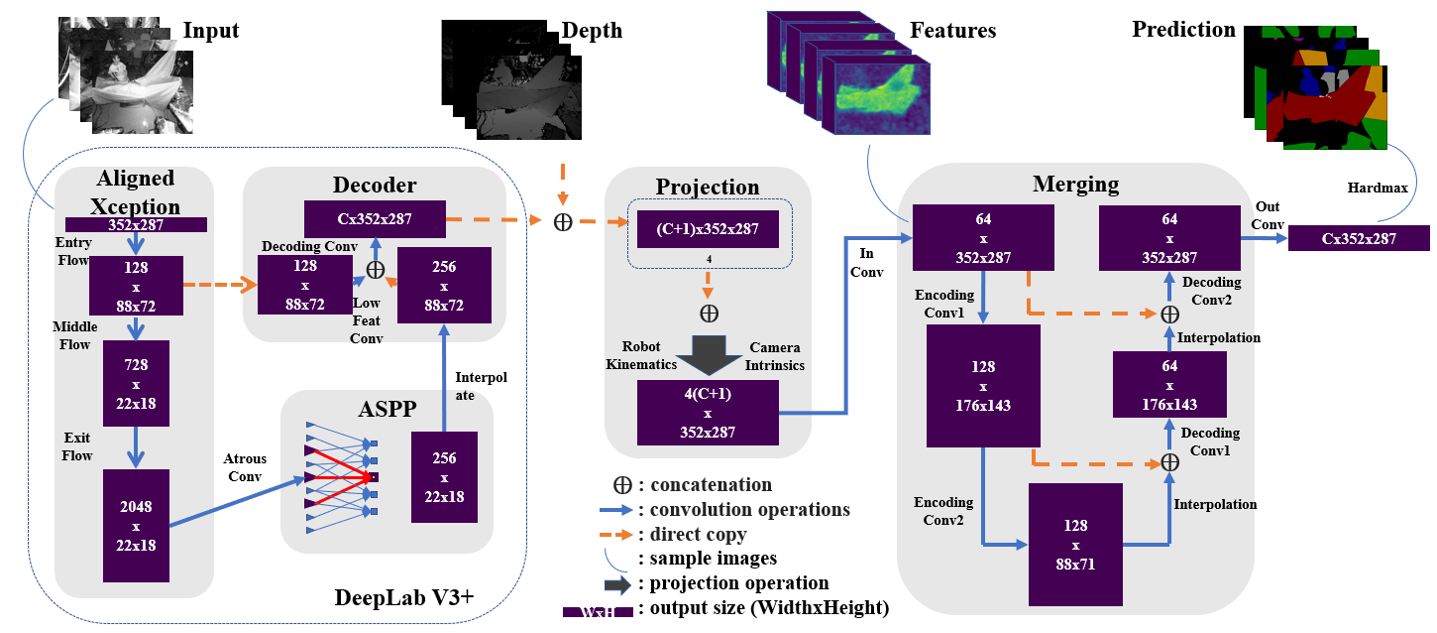}
    \caption{The MVPM model architecture. Four ToF intensity images input to the DeepLab V3+. The depth data and confidence are concatenated and projected using the robot kinematics and camera intrinsic parameters, which quadruples the information for each camera. A merging module combines the information into four segmentation result. Details can be found in the Appendix D.}
    \label{fig:workflow_mvpm}
\end{figure}

\section{Result}
\label{sec:result}
\subsection{Calibration Evaluation}
\label{sec:calibration_eval}
The sensor package is mounted on four different da Vinci Xi systems and the same calibration fixture is used to calculate the target-residual-error (TRE) by transforming the fixture into each camera frame and calculate the relative distances. The result, reported as percentage of TRE over object-camera distance, is summarized in \autoref{tab:tre}. 

\begin{table}[htpb]
    \centering
    \caption[justification=centering]{TRE for camera calibration on different da Vinci Xi systems}
    \label{tab:tre}      
    \begin{tabularx}{\textwidth}{XYYYYY}
        \hline\noalign{\smallskip}
        \textbf{System ID} &  1 & 2 & 3 & 4 & \textbf{Average} \\
        \noalign{\smallskip}\hline\noalign{\smallskip}
        \textbf{TRE\%} & $2.1\% \pm 0.5\%$ & $3.0\% \pm 1.2\%$ &  $4.1\% \pm 1.6\%$ &  $3.3\% \pm 1.3\%$ & {\pmb{$3.3\% \pm 1.4\%$}} \\
        \noalign{\smallskip}\hline
    \end{tabularx}
\end{table}

The calibration has on average $3.3\%$ error with $1\%$ error inherent from the sensor itself. 
\textcolor{black}{Additional parameterized calibration error analysis can be found in Appendix E.} One main source of error is from the inaccurate camera distortion parameters that came from the sensor manufacturer. Examples of the distortion effect can be found in Appendix F. In the future, further systematic distortion calibration is needed. 

\subsection{Semantic Segmentation Evaluation}
\label{sec:sem_seg_eval}
The training of the network has two phases: the training of the DeepLab V3+ base model and the merging model. This separate training strategy is adopted since not all data have multi-view correspondence. The 7980 single-view images are split randomly into two parts, training (80\%) and validation (20\%). 
\textcolor{black}{The multi-view data is split via a k-fold approach (k=10). In each k-fold experiment, 20\% of the data is first held out randomly as test set. The rest of the data is then partitioned into 10 folds, where 9 folds are used for training and 1 fold is used for validation. The 10-fold experiment is then repeated for 5 times, totaling 50 experiments. Each repetition of the k-fold experiment has a unique split of the data (including training, validation and test data).} 

The DeepLab V3+ is first trained on the 6384 single-view training images for 400 epochs and then validated on the 596 single-view validation images. For each k-fold experiment, to compare DeepLab V3+ and MVPM fairly, the DeepLab V3+ is fine-tuned for 50 epochs on the multi-view training split, and validated on the single-view validation split to avoid over-fitting on the small multi-view dataset. The merging module is trained for 100 epochs on the multi-view training split and validated on the multi-view split images. Details of the training parameters can be found in the Appendix G.

The MVPM approach is compared with the base-line DeepLab V3+ and Dense 3D CRF. The CRF projects the points with their features to a high dimensional space, approximates them on a permutohedral lattice \cite{adams2010fast} and performs bilateral filtering on the points unary and pairwise potentials. The unary potential of a point is the softmax probability output from the DeepLab V3+, and the pairwise potential takes both the appearance and smoothness into account. The CRF parameters used in the experiment are the default parameters used in \cite{krahenbuhl2011efficient}. 

\color{black}
The overall pixel accuracy (Acc), pixel accuracy averaged over classes (Acc\textsubscript{class}), mean Intersection Over Union (mIOU) and freqeuncy weighted Intersection Over Union (fwIOU) on the test data are summarized in \autoref{tab:segmentation_result}. Due to the limited size of the dataset, there are cases where the class distributions of training/validation/test distribution are hard to match even if stratified k-fold approach is used, especially for the less frequently appearing classes. Therefore, there is variation of the DeepLab V3+, and it carries over to the CRF and MVPM result. The MVPM shows statistically significant improvement in Acc\textsubscript{class} of 0.025 with p-value=0.010, while CRF does not show any statistically significant improvement.

\begin{table}[htpb]
    \color{black}
    \centering
    \caption[justification=centering]{Segmentation result comparing DeepLab V3+ (single view), CRF and MVPM. Asterisk sign indicates statistical significance.}
    \label{tab:segmentation_result}       
    \begin{tabular}{lllll}
    \hline\noalign{\smallskip}
     & \textbf{Acc} & \textbf{Acc\textsubscript{class}*} & \textbf{mIOU} & \textbf{fwIOU}\\
    \noalign{\smallskip}\hline\noalign{\smallskip}
    DeepLab V3+ & $0.921\pm0.008$& $0.756\pm 0.043$ & $0.670 \pm 0.036$ & $0.857\pm0.014$ \\
     CRF & $0.922\pm0.008$ & $0.754 \pm 0.044$ & $0.671 \pm 0.036$ & $0.859 \pm 0.014$ \\
    MVPM & $0.923\pm0.006$ & $0.781 \pm 0.035$ & $ 0.685 \pm 0.028$ & $0.860 \pm 0.011$\\
    \noalign{\smallskip}\hline
\end{tabular}
\end{table}

Furthermore, the per-class mIOU is computed and analyzed. The evaluated approaches have similar performance for the high frequency classes, while the MVPM statistically significantly improves the result for less frequently appearing classes, such as Human, Mayo Stand, Sterile Table and Anesthesia Cart. The MVPM approach improves mIOU results to $0.711\pm 0.031$, $0.445\pm0.034$, $0.661\pm0.030$, $0.687\pm0.060$ from baseline model mIOU of $0.697\pm0.032$, $0.425\pm0.045$, $0.646\pm0.037$, $0.650\pm0.078$ with an improvement of 0.013, 0.020, 0.015 and 0.037 for each of the above classes respectively. Other less frequently appearing classes with pixel frequency less than 1\%, such as VSC and Ceiling Light, do not have statistically significant improvement due to limited appearance. The CRF does not have any statistically significant improvement. The full result can be found in Appendix H.

\begin{figure}[htpb]
    \centering
    \includegraphics[width=0.55\linewidth]{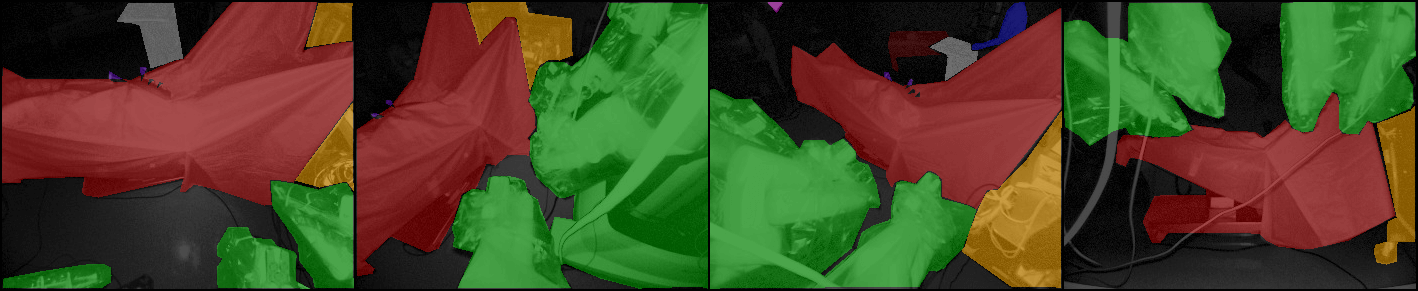}
    
    \includegraphics[width=0.55\linewidth]{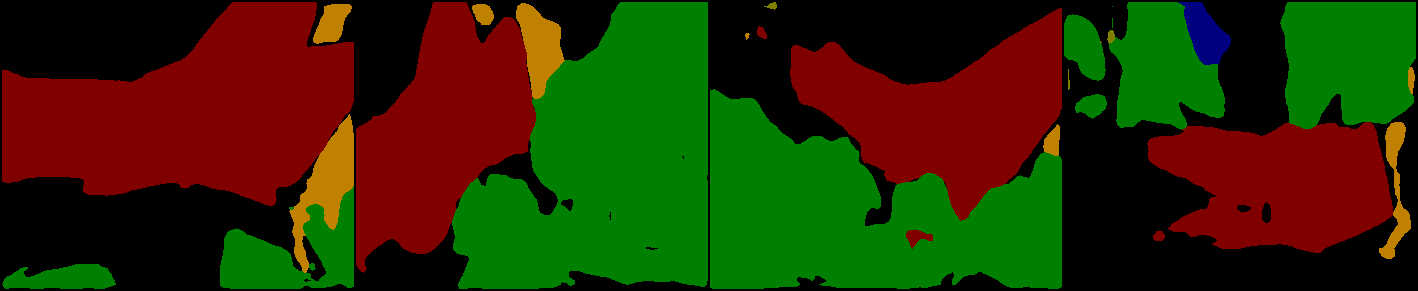}
    
    \includegraphics[width=0.55\linewidth]{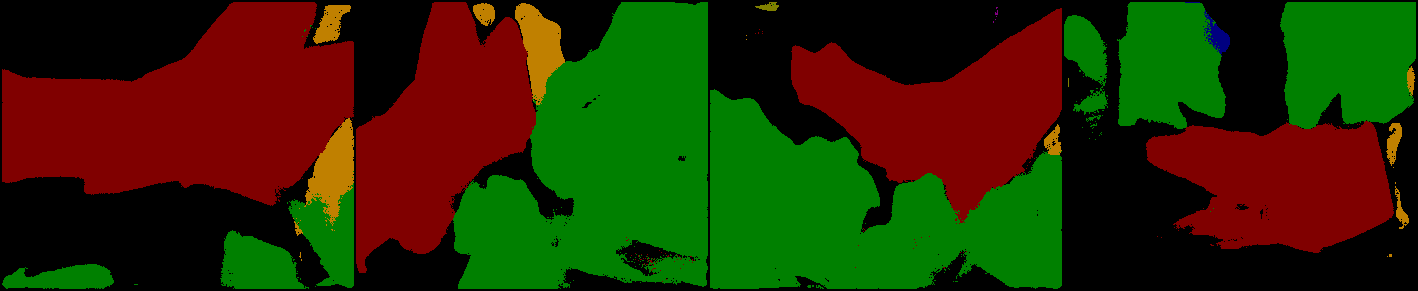}
    
    \includegraphics[width=0.55\linewidth]{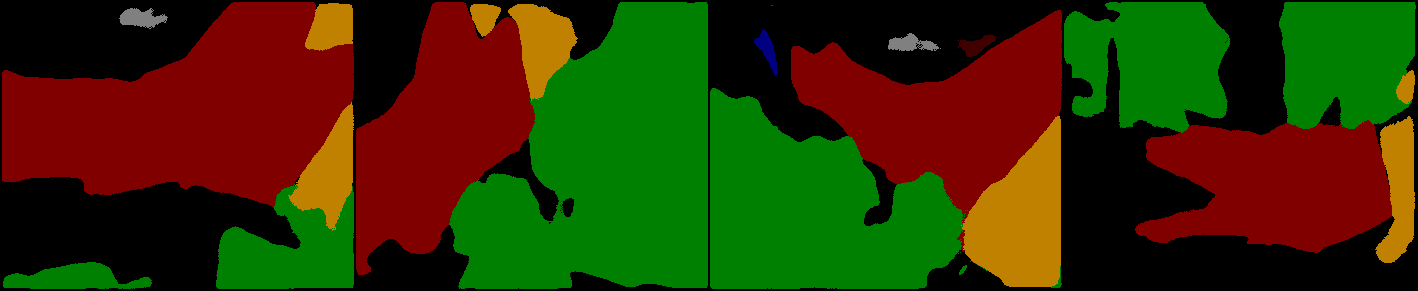}
    \caption{Example of the results in test data (grid image shows OP, USM1, USM4, BASE cameras views from left to right), (a) overlay of input and ground truth, (b) DeepLab V3+ single view prediction, (c) Dense 3D CRF, (d) MVPM prediction.}
    \label{fig:segmentation_result}
\end{figure}

\color{black}
The MVPM algorithm can enhance the pixel-pixel relationship and provide region-smoothing like the CRF approach. It also improves the prediction of the hard-to-segment objects by combining the confidence of different views in per-class manner. In the case shown in \autoref{fig:segmentation_result}, the anesthesia cart is hardly visible in all views, with the single-view prediction only capturing the object partially. The MVPM algorithm successfully merges result from different views and leads to the best performance. 

However, since the proposed calibration process in \autoref{sec:calibration} just enabled the multi-view data collection, the size of multi-view dataset at the time of submission is much smaller than single-view dataset. Therefore, in future work, the multi-view dataset will be expanded and the proposed method would be validated on a larger dataset. The imbalance of the classes should also be addressed in data collection process. Moreover, the default parameters of the Dense 3D CRF are used. In future work, the CRF parameters can be learned from the data in an end-to-end approach.

\subsection{Ablation Study}
\label{sec:ablation}
The experiment evaluates the performance of MVPM when different numbers of camera are available \textcolor{black}{using a k-fold (k=10) approach}. In this experiment, each camera view is validated by adding other cameras incrementally. \textcolor{black}{The mIOU are $0.690 \pm 0.070,0.592\pm0.061,0.491\pm0.045,0.395\pm0.038$ respectively when the camera number decreases from 4 to 1.} The best performance appears when all cameras are available \textcolor{black}{(p-value=4.11E-9)}. This proves the contribution from the mutual information in the scene. 
In future work, different view-points should be weighted based on similarities to explicitly state the preferences over camera views. Additional ablation study evaluating the effect of dropout can be found in Appendix D.4.

\color{black}
\section{Discussion}
\label{sec:discussion}
The results show that the system has an acceptable TRE of $3.3\%\pm1.4\%$. The MVPM outperforms DeepLab V3+ and Dense 3D CRF for semantic segmentation, with statistically significant improvement in less frequently appearing classes such as Human, Mayo Stand, Sterile Table and Anesthesia Cart. Ablation studies have also shown performance improvements as the number of cameras increase. However, the results are limited due to the multi-view dataset size and its distribution. Future work includes obtaining more accurate camera parameters to improve registration accuracy and working towards a larger and more balanced multi-view dataset.

\color{black}
\section{Conclusion}
\label{sec:conclusion}
In this paper, we describe for the first time, a complete and novel solution to create a 3D perception system to enhance environmental awareness, for surgical robots like the da Vinci Xi system. Our system consists of four ToF cameras rigidly mounted on the PSC robot, and a one-time calibration process for multi-camera to robot registration, which is sufficient for the sensor package to be used in other da Vinci Xi systems. Furthermore, a multi-view semantic segmentation fusion framework called MVPM and new datasets for algorithm training/validation are proposed. Our results show that the framework can improve OR scene segmentation accuracy over single camera prediction. \textcolor{black}{The proposed system can be used as a building block technology for applications such as surgical workflow analysis, automation of surgical sub-tasks and advanced guidance systems.}


%

\textbf{Conflict of interest:} Z. Li, A. Shaban and J.G. Simard are former interns, D. Rabindran, S. DiMaio and O. Mohareri are employees at Intuitive Surgical Inc.

\textbf{Ethical approval:} For this type of study formal patient consent is not required. 

\textcolor{black}{\textbf{Informed consent:} Statement of informed patient consent was not applicable since the manuscript does not contain any patient data.}

\bibliographystyle{unsrt}
\bibliography{ref.bib}   

\end{document}